\pdfoutput=1
\PassOptionsToPackage{capitalise}{cleveref}

\newif\ifarxivstyle
\arxivstyletrue

\ifarxivstyle
  \documentclass[11pt]{article}
\else
  \documentclass[
    10pt,
    accent=2F4858,
    titlefont=serif,
    abstractfont=serif,
  ]{techreport}
\fi

\ifarxivstyle\usepackage{arxivstyle}\fi
\usepackage[symbol]{footmisc}
\usepackage[most]{tcolorbox}
\usepackage{float}

\hypersetup{
    pdftitle={dig bench},
    pdfpagemode=FullScreen,
    }
\usepackage{enumitem}
\usepackage{xurl}
\newcommand{\modelid}[1]{\nolinkurl{#1}}
\usepackage{verbatim}

\newtcolorbox{playertext}[1][]{
    enhanced, breakable, parbox=false,
    colback=black!3, colframe=black!25,
    colbacktitle=black!8, coltitle=black,
    boxrule=0.4pt, titlerule=0pt, arc=2pt,
    left=10pt, right=10pt, top=9pt, bottom=9pt,
    fonttitle=\bfseries\small,
    before upper={\setlength{\parskip}{0.6em}\setlength{\parindent}{0pt}%
                  \tolerance=3000 \emergencystretch=3em \hbadness=10000},
    #1
}

\newtcolorbox{playerquote}{
    enhanced, breakable, parbox=false,
    colback=white, frame hidden,
    borderline west={1.2pt}{0pt}{black!25},
    boxrule=0pt, arc=0pt,
    left=10pt, right=4pt, top=6pt, bottom=6pt,
    before skip=12pt, after skip=12pt,
    before upper={\setlength{\parskip}{0.5em}\setlength{\parindent}{0pt}},
}

\newcommand{\widefigwidth}{1.0\textwidth}
\newcommand{\widefig}[1]{\noindent\makebox[\textwidth][c]{\includegraphics[width=\widefigwidth]{#1}}}

\newcommand{\cropfig}[2]{%
  \noindent\makebox[\textwidth][c]{%
    \includegraphics[trim={#1}, clip, width=\widefigwidth]{#2}}}

\newlength{\fracfigwidth}
\newcommand{\fracfig}[2]{%
  \setlength{\fracfigwidth}{\widefigwidth}%
  \setlength{\fracfigwidth}{#1\fracfigwidth}%
  \noindent\makebox[\textwidth][c]{\includegraphics[width=\fracfigwidth]{#2}}}

\newif\ifusealesfigureone
\usealesfigureonetrue

\newif\ifshowauthorlist
\showauthorlisttrue

\title{DiG-bench: Discovery in Games}
\ifshowauthorlist
\author[1]{Ruairidh~M.~Battleday}
\author[1]{Kai~Sandbrink}
\author[1]{Jimi~Cullen-Drohan}
\author[1]{Zihan~Yan}
\author[1]{Timothy~Muller}
\author[1]{Clare~Maguire}
\author[1]{Ales~Kubicek}
\author[1]{Fraser~Greenlee-Scott}
\author[1]{Sukrit~Sumant}
\author[3]{Tri~Dao}
\author[4, 5]{J\"urgen~Schmidhuber}
\author[6]{Michal~Valko}
\author[7]{Joshua~Tenenbaum}
\author[3]{Thomas~L.~Griffiths}
\author[2]{Zeb~Kurth-Nelson}
\author[1, 2]{James~C.R.~Whittington}
\affiliation[1]{Prefrontal}
\affiliation[2]{University of Oxford}
\affiliation[3]{Princeton University}
\affiliation[4]{King~Abdullah~University~of~Science~and~Technology}
\affiliation[5]{Swiss AI Lab}
\affiliation[6]{Inria}
\affiliation[7]{Massachusetts~Institute~of~Technology}

\else

\author{XXX}
\fi

\abstract{
Discovery---formulating novel generalizations---is a central part of the scientific process. Despite its importance, there is a gap in the current AI benchmark landscape, with few benchmarks directly probing the capacity for discovering new knowledge with experimentation in controlled environments where the objective is unknown. To address this gap, we release a new benchmark: DiG-bench (Discovery in Games). DiG-bench consists of a set of 70 independent games. Each game is encoded as a short string and has unique transformation rules that must be discovered through interaction and experimentation. The levels of the game present a series of challenges to test whether the rules have been discovered, where the win conditions for each level are also unknown. We provide games at seven tiers of difficulty for AI agents. The lowest tier is routinely solvable by multiple models, while the highest tier challenges the best models in agentic harnesses. All 70 games were solved by at least one human on first attempt. A subset of 21 games is released publicly, and the remainder is held private for secure evaluation.
\par\medskip
Benchmark website: \url{https://digbench.ai}
}

\begin{document}
\ifarxivstyle\else\vspace*{-1.0cm}\fi
\maketitle
\ifarxivstyle\else\vspace{-0.4cm}\fi

\begin{figure}[H]
    \centering
    \ifusealesfigureone
        \widefig{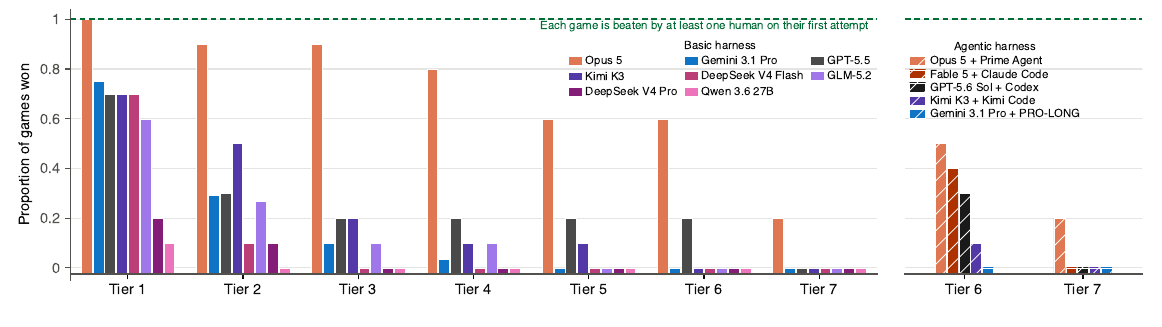}
        \caption{\textbf{Model performance on DiG-bench.} Proportion of games won by each model in each tier. Models in the basic harness were evaluated on all seven tiers (left). Models in agentic harnesses were evaluated only on tiers 6 and 7 (right). The horizontal dashed line indicates that humans beat every game.}
        \label{fig:results}
    \else
        \widefig{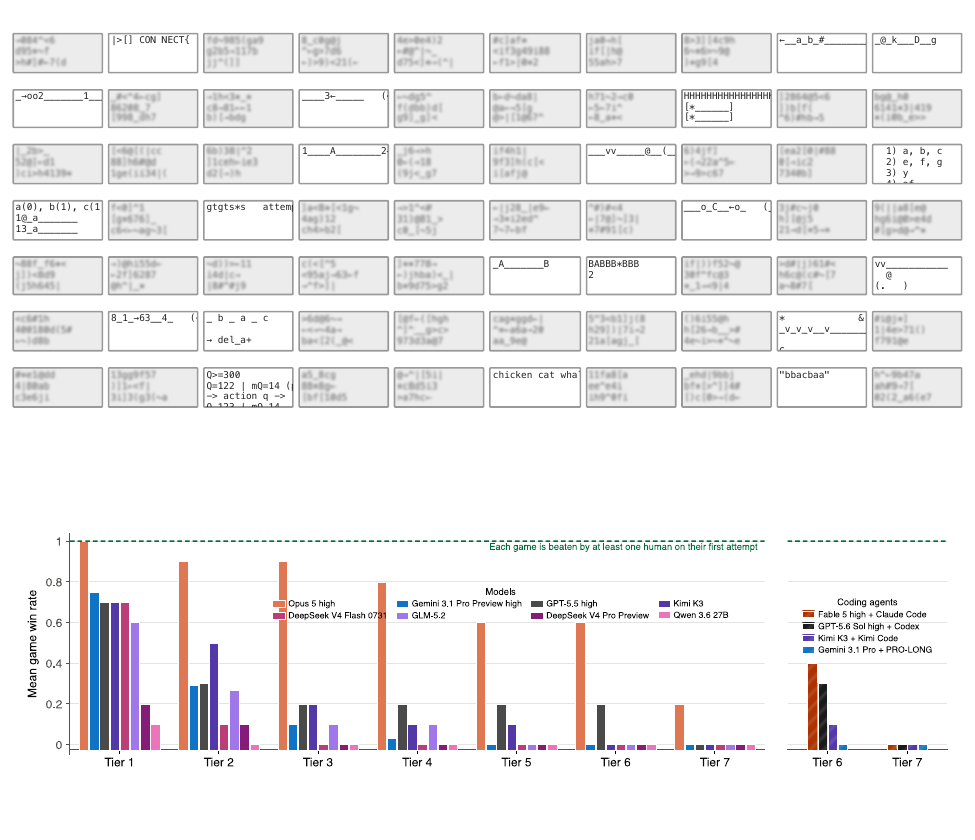}
        \caption{\textbf{The DiG-bench benchmark.} Top: DiG-bench consists of 70 games, of which 21 are public and 49 are private, organized into seven tiers of 10 games each. Bottom: mean game win rate grouped by tier, with basic harness conditions across all seven tiers (left) and agentic harness conditions for tiers 6 and 7 (right). Binary outcomes are averaged within each model--game pair before the 10 game scores are averaged with equal weight; an em dash marks no eligible run (see \cref{sec:llm-types}). The horizontal dashed line indicates that humans beat every game.}
        \label{fig:results}\label{fig:interface}
    \fi
\end{figure}

\ifusealesfigureone
\begin{figure}[H]
    \centering
    \widefig{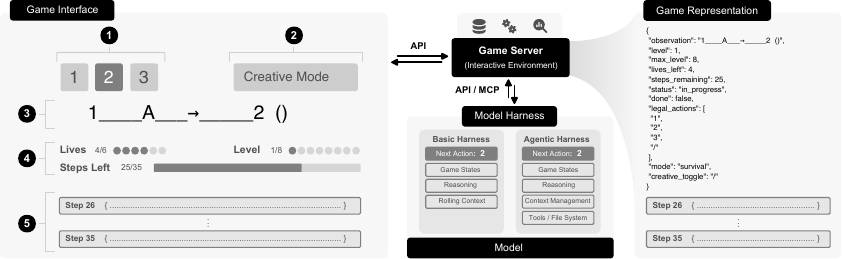}
    \caption{\textbf{The DiG-bench platform.} The platform and equivalent human/model interfaces. \includegraphics[scale=0.9,trim=0 1 0 0]{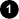} the available actions (action `2' is highlighted because the player is selecting it); \includegraphics[scale=0.9,trim=0 1 0 0]{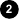} creative mode, an option in some games that allows a player to enter a sandbox for experimentation; \includegraphics[scale=0.9,trim=0 1 0 0]{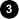} the current game observation; \includegraphics[scale=0.9,trim=0 1 0 0]{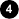} indicators of current level, remaining lives and remaining steps; and \includegraphics[scale=0.9,trim=0 1 0 0]{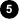} action/state history.}
    \label{fig:interface}
\end{figure}
\fi
\section{Introduction}

Discovering new knowledge drives science forward. We define discovery as finding and making sense of previously unexplained regularities in a way that is useful for compressing existing observations \cite{schmidhuber2008driven} and reasoning about new states of a system. There is now enormous interest in building AI systems that can accelerate this process, from AI research to structural biology, chemistry, and mathematics~\cite{andrej_karpathyautoresearch_2026, ghareeb.etal2026}.

Measuring progress toward AI that can make genuine discoveries requires a benchmark that isolates the capacity for discovery itself. Some existing benchmarks, like ARC-AGI-3~\cite{arc_prize_foundation_arc-agi-3_2026}, are framed as tests of discovery and fluid reasoning, but confound this with visual perception. Others, like ARC-AGI-1/2~\cite{chollet_arc_2025, chollet_arc-agi-2_2025}, Bongard-LOGO~\cite{nie_bongard-logo_2020} and IOLBench~\cite{goyal_iolbench_2025}, probe rule induction but not active experimentation. Yet others, such as DiscoveryWorld~\cite{jansen_discoveryworld_2024}, test experimentation but entangle discovery with prior scientific knowledge or rely on a `menu' of discoveries. Taken together, this means the current benchmark landscape lacks a targeted, controlled benchmark for active discovery.

We therefore introduce DiG-bench (Discovery in Games), a benchmark of 70 games designed to map the surface of discovery in well-controlled interactive systems. Games have a long history in AI~\cite{yannakakis2018artificial}. In DiG-bench, each game is a self-contained miniature world with its own laws, but both the rules and the objective are hidden from the player and must be uncovered through interaction. The games exercise the discovery process end-to-end, with rich possibilities for experimentation.

Six design choices distinguish DiG-bench, and each follows directly from the goal of measuring discovery in isolation. 

\begin{enumerate}
    \item The games operate within the \textbf{natural domain of large language models}, creating the best test of the models' true discovery capabilities. The games are purely text-based: observations are short strings, usually on a single line.
    This is done so that there are no visuospatial confounds~\cite{wang_your_2026}, leaving discovery as the operative challenge. Games are short enough that most traces fit entirely within the context window of current frontier models, so that we test for discovery ``in-context'' without any need for weight updates or complicated context management.
    \item The games are \textbf{handcrafted} by human experts and \textbf{novel}, and we keep the majority of them private.
    \item We acknowledge that \textbf{discovery is effortful}. All games have been solved by at least one human player on that player's first exposure to the game---but players reported finding many games difficult, and traces show extensive experimentation.
    \item Discoveries across games reflect a \textbf{rich and diverse set of mechanisms}, meaning that the benchmark is not solved by solving a single challenge, such as vision.  
    \item \textbf{Experimentation} is a central part of the discovery process. For this, many games contain a special creative mode that allows players to test mechanics in a sandbox-like environment with a less restrictive step limit. Within a game, the ability to perform informative experiments often depends on understanding previous experiences.
    \item Difficulty is \textbf{calibrated to the frontier}. We find tasks that are right at the tipping point of what models can do, allowing us to map the surface of discovery accurately.
\end{enumerate}

The remainder of this report is organized as follows. We first outline the benchmark~(\cref{sec:benchmark}), then report the performance of humans and LLM agents and its breakdown across difficulty tiers~(\cref{sec:results}). We perform a separate analysis in which we give Gemini 3.1 Pro access to the true rules of each game, to evaluate how much easier the games are if the rules are known. We then detail the gameplay setup~(\cref{sec:methods}), situate the benchmark against related work~(\cref{sec:comparison}), and close with a discussion of the skills needed to solve the games and directions for future work~(\cref{sec:discussion}). Appendices present prompts and further analyses.

\begin{figure}[H]
    \centering
    \cropfig{58 3.5 57 4}{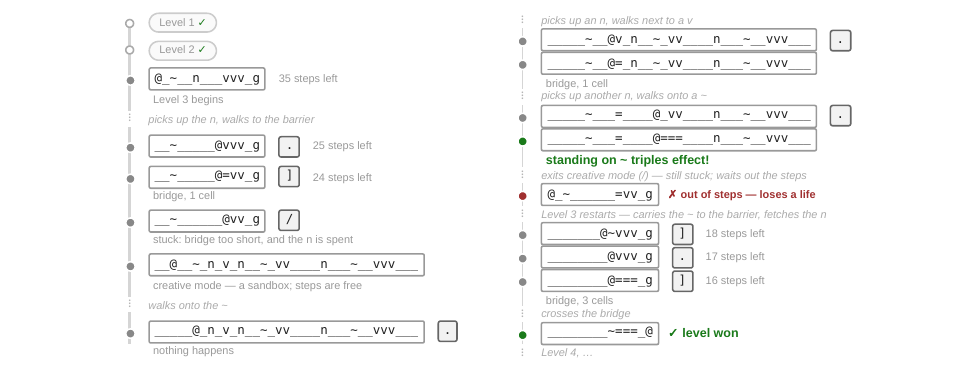}
    \caption{\textbf{Gameplay in game P-21.} An abridged example of a gameplay timeline. Having beaten Levels 1 and 2, the player applies the rule learned on Level 2 (not shown): activating a held \texttt{n} with `\texttt{.}' builds a bridge. But Level 3 contains a triple barrier \texttt{vvv}, and the bridge is too short. The player enters creative mode, a sandbox with a less restrictive step limit. Experimenting, the player discovers that standing on \texttt{\textasciitilde} has a tripling effect, allowing a longer bridge to be built. Back in survival mode, the level can no longer be won, so the player waits out the remaining steps to restart Level 3. On the second attempt, the player carries the \texttt{\textasciitilde} to the barrier, activates the \texttt{n} while standing on \texttt{\textasciitilde}, and crosses the triple-length bridge to reach the goal \texttt{g}, winning the level.}
    \label{fig:sequence}
\end{figure}

\begin{table}[H]
    \centering
    \caption{\textbf{Examples of discoveries in public games.} Each row summarizes one mechanic a player must uncover to solve the game.}
    \label{tab:discovery-examples}
    \begingroup
    \small
    \renewcommand{\arraystretch}{1.25}
    \begin{tabular}{@{}p{0.13\textwidth}p{0.79\textwidth}@{}}
        \toprule
        \textbf{Game} & \textbf{Example discovery} \\
        \midrule
        P-2  & Every level has two parts with analogous structure between the two parts. \\
        P-8  & Lilies grow only on the graves of roses. \\
        P-9  & Stepping over an underscore reverses the goal sequence order. \\
        P-13 & When Knights outnumber Knaves, Knaves tell the truth. \\
        P-14 & Voles run away from beetles. \\
        P-15 & Reactions have both parallel and rotationally symmetric structure. \\
        P-18 & The sliders control rate constants. \\
        P-20 & The semantic meaning of each word is informative about how the brackets operate on it. \\
        \bottomrule
    \end{tabular}
    \endgroup
\end{table}

\section{Benchmark}

\label{sec:benchmark}

The benchmark contains 70 games. We assigned each game to one of seven tiers according to machine difficulty, with tier 1 being the easiest and tier 7 the hardest~(\cref{fig:results}). The lowest tier is mostly beaten by one-generation-old models like Gemini 3.1 Pro, while the higher tiers are challenging for state of the art models, including those in agentic harnesses. All 70 games were beaten by at least one human on their first attempt. We release three games publicly for each of the seven tiers, for a total of 21 public games, named P-1 through P-21. We hold the remaining games private for evaluation.

\begin{figure}
    \centering
    \widefig{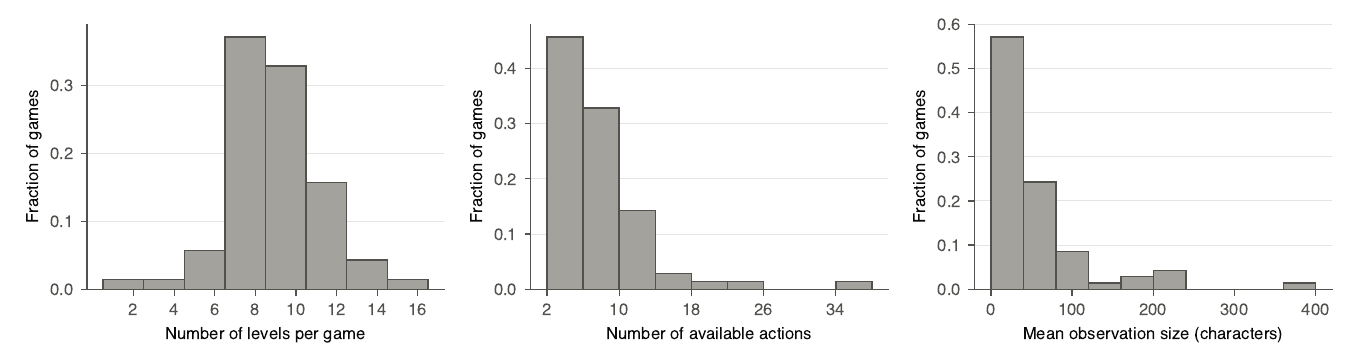}
    \caption{\textbf{Descriptive statistics of games.} Distribution over all 70 games of the number of levels per game (left), the number of available actions (center), and the mean observation size (right).}
    \label{fig:design-stats}
\end{figure}

At each step of a game, the game generates an observation based on the current state, the player sees the observation and responds with an action, and the action conditions the next state transition in the game~(\cref{fig:interface,fig:sequence}). The games thus have the form of a partially observable Markov decision process (POMDP). In some games, multiple states alias into the same observation, while other games are fully observable. In our benchmark, observations are text-based and formatted as a short Unicode string. Most games' observations consist of letters, numbers and simple punctuation; a few use special characters like arrows. Many observations fall on a single line; others include a small number of newline characters, for example to show an inventory or to place a cursor beneath a line. Newlines are not used to build large 2D grids. Actions correspond to single characters. Most games have fewer than 10 total possible actions
(\cref{fig:design-stats}). Which actions are available, and what effect they have, can change dynamically within a game. Each game has between 1 and 16 levels through which players progressively discover the game's structure, and each level has a limited number of steps~(\cref{fig:design-stats}). 

To facilitate experimentation, many games include a creative mode~(\cref{fig:sequence}) that can be entered and exited via a special action corresponding to a forward slash ``/''. Creative mode is a sandbox with rules similar or identical to the main game, but where steps do not count towards the per-level limit, allowing free exploration (there is a large, but finite, separate limit for steps taken in creative mode). The initial state in creative mode is typically different from the main game, so the player cannot simply work out solutions and copy them. 

While playing the game, both humans and models are given access to the complete observation-action history in the game (although some model-harness combinations experienced context truncation in a small number of games). 

\begin{figure}
    \centering
    \fracfig{0.8}{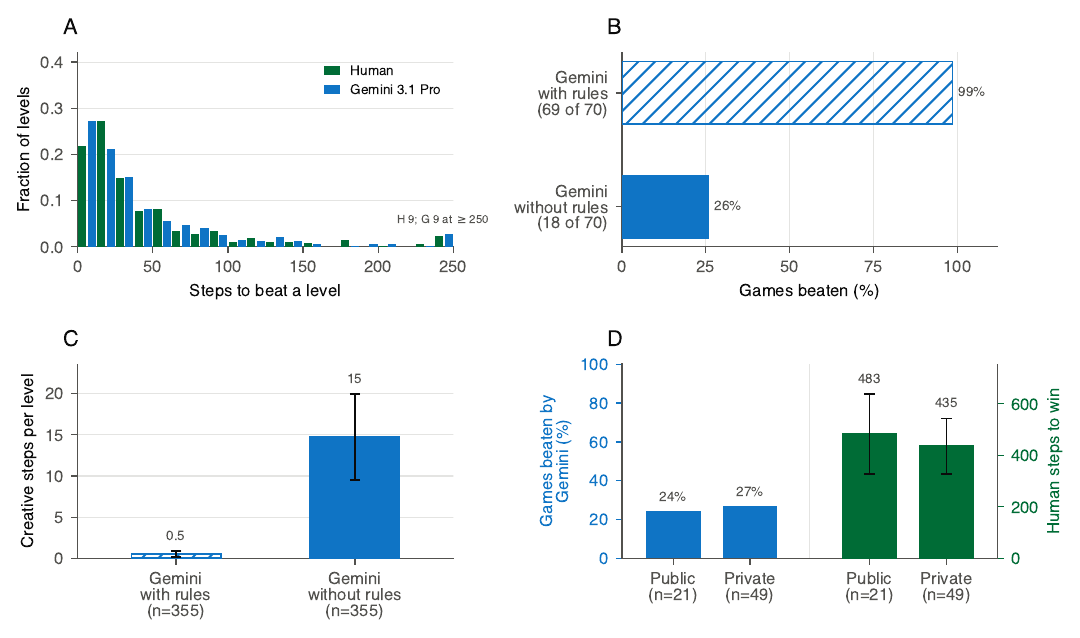}
    \caption{\textbf{Gameplay data.}
    (\textbf{A}) Steps to beat a level by humans versus Gemini 3.1 Pro, for all the levels beaten by both; the final bin includes 250 steps or more. (\textbf{B}) Giving Gemini access to the ground-truth rules of the game, specified in concise natural language, raises the win rate from 18/70 to 69/70 games. (\textbf{C}) Creative-mode steps per level by Gemini with and without access to rules. Error bars are 95\% confidence intervals on the mean. (\textbf{D}) Left: Gemini win rates are similar between public and private games. Right: human step efficiency is similar between public and private games.}
    \label{fig:design-perf}
\end{figure}

\FloatBarrier

\section{Results}
\label{sec:results}

We evaluated a range of AI models against our benchmark. DeepSeek V4 Pro Preview, Gemini 3.1 Pro, Opus 5, Kimi K3, DeepSeek V4 Flash 0731, Qwen 3.6 27B, GPT 5.5 and GLM 5.2 were evaluated on all games. Results are shown in \cref{fig:results}. The weakest model, Qwen 3.6 27B, beat 1 out of 70 games. The strongest model, Opus 5, beat 50. Taking the best model separately for each game after forming those model$\times$game means, models in the basic harness collectively scored 57 out of 70; within the top two tiers, they scored 9 out of 20.

We wanted to know whether models in an agentic harness would outperform models in the basic harness. We tested Fable 5 in Claude Code, GPT 5.6 Sol in Codex, Kimi K3 in Kimi Code, and Gemini 3.1 Pro in PRO-LONG. The agentic harness conditions were tested only on tiers 6 and 7. In the case of Kimi K3 and Gemini 3.1 Pro, where a direct comparison was available, the model in the agentic harness performed no better than the same model in the basic harness. Finally, we tested Prime Agent (Opus 5 running in the Prime Agent product), which recently achieved 95.5\% on ARC-AGI-3 public games, against tiers 6 and 7 of our benchmark. Prime Agent did not improve over basic harness Opus 5.

One aim of our benchmark is to test the player's capacity for deeper experimentation: constructing informative situations that are not directly along the path to a goal. Therefore, unlike ARC-AGI-3, our benchmark does not score step efficiency: exploration that stays within the step limit is not penalized. Of the levels beaten by both, Gemini 3.1 Pro took 46 $\pm$ 63 steps per level and humans took 49 $\pm$ 78 (mean $\pm$ SD; game-level paired Wilcoxon $p=0.15$; see \cref{fig:design-perf}A).

If the games truly challenge discovery rather than other abilities like planning or long-context retrieval, then \textit{telling} models the rules of the game should categorically improve their performance. We therefore reran Gemini 3.1 Pro under the same conditions as before, but supplied with an additional text field which was a compact natural language description of the rules of the game. These rules summarized the dynamics of the game and the win condition; but they excluded any policy-related information like strategies, tactics or move sequences. Gemini beat 69 out of the 70 games with the rules given, against 18 out of 70 without them (\cref{fig:design-perf}B). On the one game that Gemini did not beat with the rules, access to the rules nonetheless markedly improved its performance~(\cref{app:unbeaten-with-rules}). Furthermore, giving Gemini the rules almost eliminated its use of creative mode (\cref{fig:design-perf}C). These data are consistent with the idea that a primary challenge of the benchmark is finding out the rules. 

Finally, we had humans play the games to ensure that every game was human beatable. All 70 games were beaten by at least one human on their first attempt at that game. Humans reported finding the games challenging, with some plays lasting for more than an hour of continuous play. Some players reported using a pen and paper. Our favorite quote from the web feedback form we gave players to fill out after they had finished a game: `during dinner I was trying to figure out how to solve the puzzles'. The number of steps used by humans was no different than Gemini 3.1 Pro in matched levels (\cref{fig:design-perf}A).

\section{Methods}
\label{sec:methods}

\subsection{Agents}
\label{sec:agents}

LLM agents interacted with the benchmark via an API that we are making publicly available for the public games. Agents were told that each game has a certain number of levels that they need to complete within a certain number of lives, and that they need to complete each level within a certain number of steps. Agents received the same information as humans as a prompt containing a general task description (see \cref{app:instructions}), the current observation, level, lives, remaining steps, status, and legal actions, plus a game-mode field and level-transition message where applicable as part of their prompt. They then select an action using a tool call or in their output. We describe the agent harnesses we used in \cref{sec:agent-harnesses}. We evaluated a broad set of models, with coverage varying by model and tier. Closed-source models were run using official provider or AWS Bedrock APIs; some models were run with a cost cap listed in \cref{tab:model-config}. Open-source models were run on a hosted server using NVIDIA H200 GPUs (on their high reasoning effort setting where available). Most model$\times$game pairs had only a single run, so the results are effectively single-seed and we do not report variability across seeds. In some instances a model played a game more than once (with fresh context) due to evolving research plans. In Figure 1, we average those runs, because the aim is a fair comparison between models. In Figure 5 and 7, we count a game as beaten by Gemini 3.1 Pro if any Gemini run on that game resulted in a win, because we are interested in whether the games are beatable.

\paragraph{Model identifiers} The exact model identifiers recorded for these results are \modelid{global.anthropic.claude-opus-5}, \modelid{openai.gpt-5.5}, \modelid{gemini-3.1-pro-preview}, \modelid{kimi-k3}, \modelid{glm-5.2}, \modelid{deepseek-v4-pro}, \modelid{deepseek-v4-flash}, \modelid{qwen3.6-27b}, \modelid{global.anthropic.claude-fable-5}, and \modelid{openai.gpt-5.6-sol}. The Anthropic and OpenAI models were accessed through Amazon Bedrock; Gemini 3.1 Pro was accessed through Google's own API. Because they were served through Bedrock, the OpenAI models were limited to a 272k-token context window rather than the 1M they support natively. For open-weights models the identifier is the checkpoint we served rather than an API identifier. Where a reasoning-effort setting was exposed, it was set to \texttt{high}. The runs reported were collected between 13 May and 11 August 2026; human plays were collected between 23 June and 5 August 2026.

\paragraph{Runs stopped at a cap} A run that reaches its per-game cost or wall-clock cap is counted as a finished attempt that did not beat the game: a capped run is a loss. Most runs used a \$200 cost cap. Kimi K3 used a 262k-token context window---rather than the 1M it supports, for speed and cost---and a 12-hour wall-clock cap per game. GLM-5.2, DeepSeek V4 Pro Preview, DeepSeek V4 Flash 0731, and Qwen3.6 27B likewise used a 12-hour wall-clock cap per game.

\subsubsection{Harnesses}
\label{sec:agent-harnesses}

We evaluated models in one of two conditions: a minimal \emph{basic harness} or an \emph{agentic harness}, which layers a general-purpose agent product on top of the same game interface.

\paragraph{Basic harness}

The basic harness tests the ability of base LLMs to make discoveries without additional scaffolding. It preserves each provider's native reasoning-continuity mechanism where available, such as Gemini thought signatures, Anthropic thinking signatures, OpenAI encrypted reasoning items, or visible reasoning in SGLang-served open models. This lets models carry their own reasoning state across turns, giving models the best chance of performing well~\cite{bigio_enabling_2026}.

At the start of each run, the harness sends the model the task description and initial game state. It then executes a simple loop: the updated game state is appended to the conversation, the model submits exactly one legal action, and the action is applied, repeating until the game ends.

In very long games, the observation-action history is truncated to fit within the context window, excluding the oldest steps. This occurred in 6.1\% of runs, mostly Qwen3.6 27B and GPT-5.5. For GPT-5.5, truncation occurred because the model as hosted on Amazon Bedrock had a 272k-token context window rather than the model's 1M-token context window. No Gemini 3.1 Pro or Opus 5 runs had truncated context. However, we did observe anecdotally that Gemini's performance appeared to degrade as the context length grew.

\paragraph{Agentic harness}

The agentic harness conditions reported in \cref{fig:results} use Claude Code with Fable 5, Codex with GPT-5.6 Sol, Kimi Code with Kimi K3, Prime Agent \cite{primeintellect2026primeagent} with Opus 5, and PRO-LONG~\cite{fox_prolong_2026} (a Read-Grep-Bash agent based on OpenCode scaffolding) 
with Gemini 3.1 Pro. In the harnesses wrapping Claude Code, Codex and Kimi Code, the game was exposed through a Model Context Protocol (MCP) server. Prime Agent has no MCP support, so it received the same interface as a Python module preloaded into its IPython kernel. The PRO-LONG runs predate our MCP server, so PRO-LONG received the same Python-module interface as Prime Agent.
Each run executed in a Docker sandbox whose only access to the game was the interface described above, and it was scoped to a single running session so the agents could not cheat by opening a new session. Within the sandbox, the agents had access to their full native tool set. For agentic harnesses, context was managed by the harness itself.

\begin{table}[ht]
\centering
\resizebox{\linewidth}{!}{%
\begin{tabular}{lllll}
\toprule
\textbf{Harness} & \textbf{Model} & \textbf{Effort} & \textbf{Budget Cap} & \textbf{Context Size} \\
\midrule
basic harness & Claude Opus 5 & high & \$200 & 1M \\
basic harness & Gemini 3.1 Pro Preview & high & \$200 & 1M \\
basic harness & GPT-5.5 & high & \$100 & 272k \\
basic harness & Kimi K3 & max & 12-hour & 262k \\
basic harness & GLM-5.2 & max & 12-hour & 1M \\
basic harness & DeepSeek V4 Pro Preview & - & 12-hour & 1M \\
basic harness & DeepSeek V4 Flash 0731 & - & 12-hour & 1M \\
basic harness & Qwen3.6 27B & - & 12-hour & 262k\\
\midrule
agentic harness (Prime Agent) & Claude Opus 5 & high & \$200 & 1M \\
agentic harness (Claude Code) & Claude Fable 5 & high & \$200 & 1M \\
agentic harness (Codex) & GPT-5.6 Sol & high & \$200 & 272k \\
agentic harness (Kimi Code) & Kimi K3 & max & 12-hour & 1M \\
agentic harness (PRO-LONG) & Gemini 3.1 Pro Preview & high & \$200 & 1M \\
\bottomrule
\end{tabular}
}
\caption{Model Configuration}
\label{tab:model-config}
\end{table}

\subsection{Humans}
\label{sec:humans}

We collected human gold standard solutions on the games. Humans played the game via a web interface on a private portal (shown in \cref{fig:interface}). Players were invited through personal and professional networks, direct outreach to academics, requests for professors to nominate students, and puzzle-solving communities. The players tended to be motivated by solving puzzles. These game plays were used simply to confirm that the games could be beaten by humans rather than gaining generalizable insight into human behavior. All data is reported only in aggregate and anonymized form.

The gameplay experience was standardized between humans and AI agents to the greatest extent possible. Humans were given the same information as agents, except additionally instructed that they should take their time, and that they were encouraged to use a paper and pen to write down their thoughts (see \cref{app:instructions}). The rules of the game, including the goal, were not provided.  

The current observation appeared in a central box. Players could select actions either via keypress or by clicking on one of the buttons appearing above the central box. The interface also showed remaining lives, remaining steps (which reset on every level completion or lost life) and current level. The history was rendered under the interface and was accessible throughout the entire game (abridged in \cref{fig:interface}). The interface allowed humans to play each game multiple times. After finishing a game, players were asked to describe in free text how they thought it worked. Human performance is reported only from first attempts. We count a play as a first attempt only if that player had zero steps of prior exposure to the game.

\section{Related works}
\label{sec:comparison}

\paragraph{Interactive discovery environments}
\label{sec:related-interactive}

A number of benchmarks require an agent to uncover hidden task structure. Prominently, the third installment of the Abstraction and Reasoning Corpus for Artificial General Intelligence benchmark series (ARC-AGI-3) also presents games without instructions that require the agent to explore, plan, and discover the rules through interaction~\cite{arc_prize_foundation_arc-agi-3_2026}. However, these games are intentionally grounded in human spatial and visual priors for difficulty, representing the environment visually in a two-dimensional grid. This means that part of the difficulty is perceptual, with simpler underlying abstract concepts than those in our benchmark. To compare the ARC-AGI-3 games with our games more accurately, we translated five of the games into our text-based framework, each of which was beaten by Gemini 3.1 Pro in a similar number of steps as humans (\cref{app:arc-results}). Other benchmarks that test the ability of agents to beat games in visuospatial environments include the BALROG (Benchmarking Agentic LLM and VLM Reasoning On Games) agentic-reasoning suite, which features games like Baba Is AI, MiniHack, and NetHack that require deducing environmental mechanics~\cite{paglieri_balrog_2025}, and the AI GameStore~\cite{ying_ai_2026}, which aims at evaluating a model's general intelligence by adapting a set of a hundred human games. Although some of these games contain similar elements of mechanism and discovery, they explain many if not all of the rules and contain additional confounds like visual perception (in many games), long-term planning, reasoning beyond a single context window, spatial reasoning, and navigation. This makes it hard to attribute differences in performance to a single factor. Finally, benchmarks like FALSIFYBENCH~\cite{bertolazzi_falsifybench_2026}, in which agents need to discover the semantic properties of a target by proposing test cases that not only confirm but also disconfirm hypotheses, study the ability of LLMs to make inferences in abstract domains. Similarly, \citet{geng_are_2025} ask agents to reverse-engineer black-box programs, formal languages and equations, and find that active intervention helps only patchily.

\paragraph{Scientific discovery environments}
\label{sec:related-science}

Another set of interactive environments casts discovery as science, asking the agent to run experiments against a simulated world. ScienceWorld~\cite{wang_scienceworld_2022} places agents in a text environment in which they must design grounded experiments based on an elementary-school science curriculum. DiscoveryWorld~\cite{jansen_discoveryworld_2024} extends this to include what the authors call the complete cycle of novel scientific discovery, including hypothesis generation, experimental design, execution, and analysis. SciGym~\cite{duan_measuring_2025} simulates biological systems, asking agents to iteratively design experiments against them. A second group tests how efficiently an agent can probe for the hidden structure of a parametric law or model. BoxingGym~\cite{gandhi_boxinggym_2025} draws generative probabilistic models from domains like psychology and ecology; PhysGym~\cite{chen_physgym_2025} controls the amount of prior knowledge supplied in interactive physics problems; and NewtonBench~\cite{zheng_newtonbench_2025} applies counterfactual shifts to canonical physical laws to prevent recall of memorized information. CausalGame~\cite{chen_causalgame_2026} hides a structural causal model behind biased and confounded observations.

\paragraph{Rule induction from fixed demonstrations}
\label{sec:related-induction}

A final family of benchmarks tests rule inference from a fixed set of demonstrations. ARC-AGI-1 and ARC-AGI-2 present a few input--output grid examples that demonstrate some hidden transformation rule, and the system must infer that rule from the handful of examples and apply it to a new input~\cite{chollet_measure_2019, chollet_arc_2025, chollet_arc-agi-2_2025}. ConceptARC~\cite{moskvichev_conceptarc_2023} uses this format to target single spatial or semantic concepts at varying levels of abstraction to evaluate models' understanding of them more completely. ARC-AGI 1 and 2 drew on classic psychometric tests like Raven's Progressive Matrices, which test human ability to infer abstract attribute rules by asking them to complete a matrix of figures based on existing examples. Procedurally Generated Matrices~\cite{zhang_raven_2019} directly adapted this format for machine learning. Bongard-LOGO~\cite{nie_bongard-logo_2020} instead supplies a handful of positive and negative examples, and the Compositional Visual Relations (CVR) benchmark~\cite{zerroug_benchmark_2022} measures sample efficiency and compositional transfer across abstract rules. Other benchmarks remove the visual component to focus on symbolic manipulation, such as rule induction in letter-string analogies~\cite{webb_emergent_2023} or grammatical constructions~\cite{goyal_iolbench_2025}. 
This class of benchmarks isolates rule abstraction cleanly, but it cannot measure how well an agent chooses what to try next, which is the ability our benchmark is built around.

\section{Discussion}
\label{sec:discussion}

We introduce DiG-bench to measure a central capability for scientific progress: discovering new knowledge through active experimentation. DiG-bench consists of 70 new games, which do not exist anywhere on the internet, across seven tiers of difficulty. The games are designed with unfamiliar, non-obvious mechanisms discovered through interaction and experimentation. The observation of each game consists of a short string.

We tested the games on a variety of different models, either with the basic harness or with agentic harnesses. We found that many games are unbeatable by agents, even though the games are based in text---the natural domain of language models. The agentic harness conditions, including Prime Agent, did not boost performance over the basic harness. These results are suggestive that existing harnesses offer limited gains in discovery ability.

All 70 games were beaten by at least one human on a first attempt, suggesting that all of the games are beatable without excessive computation or access to unknowable information. At the same time, many of the games are substantially effortful for humans---discovery often takes work. Giving Gemini access to the ground truth rules of each game in natural language dramatically improved its performance, consistent with the idea that a primary challenge of the games is finding out the rules.

Games embedded in short strings give the best opportunity to probe the real discovery capabilities of LLMs. We found that the latest generation of models can experiment to discover unknown rules in many kinds of games created de novo. We speculate this is due to the trend in frontier models toward including a vast array of RL environments in training; therefore many short-horizon games effectively lie at points that can be interpolated from training data. Of course, discovering the rules of a game is only one kind of discovery---the most general form includes discovery of all possible kinds of knowledge.

Scientific discoveries made using AI could potentially improve human lives, ecosystems and social systems enormously. Fulfilling this potential to further science requires developing AI's ability to discover new knowledge and make open-ended discoveries in the real world. The same capability is also one of the few remaining bottlenecks to autonomous recursive self-improvement and potentially dangerous misuse; therefore, it needs to be evaluated carefully to ensure safety.  Finally, current progress in AI and related fields represents an opportunity to renew a study of human discovery, improving our understanding of human science, and potentially providing ways we can improve our own ability to make discoveries. 

\bibliographystyle{plainnat}
\bibliography{references}

\clearpage

\beginappendix
\crefalias{section}{appendix}

\section{Game instructions}
\label{app:instructions}

Humans and agents alike were told that the rules must be discovered, and were given the level, life, and step structure, without being told anything about the mechanics of any individual game (except for in the special runs where Gemini agents were given access to the ground truth rules of the game, which are shown in \cref{fig:design-perf}). The texts below reproduce the standing prompt and first turn texts given to the basic harness, the prompt template given to the agentic harness, and the instructions for human players.

Fields in braces are filled in per run: the session and game identifiers (\texttt{\{session\_id\}}, \texttt{\{game\}}), the names of the tools it is given (\texttt{\{tools\}}), its starting and first move indices (\texttt{\{step\_index\}}, \texttt{\{first\_step\}}), the requested pace (\texttt{\{pace\}}), and the observation that is currently shown (\texttt{\{state\}}). Angle brackets, by contrast, are part of the prompt text itself, telling the agent what to substitute at call time.

\begin{playertext}[title={Onboarding text shown to human players}]
We are not going to tell you the rules of this game---you have to figure them out for yourself.

These games are not easy! It may take quite a bit of thinking and tinkering to figure out what's going on. Initially, it won't make any sense at all. This is normal.

We recommend you use a paper and pen to note down your thoughts and workings.

\medskip
\textbf{Levels, lives and steps}\nopagebreak

The aim is to complete all the levels.

You advance levels by reaching certain states within the game. You will have to figure out what these are.

Within each level, you have a limited number of steps. If you run out of steps, you lose a life. If you lose all your lives, the game is over.

It is also possible to lose a life by reaching certain states within the game.

\medskip
\textbf{Important: creative mode}\nopagebreak

At any time, you can use a button to switch into ``creative mode'', where you can experiment safely without losing steps or lives.

It may be necessary to use creative mode in order to discover the rules of the game without running out of steps.
\end{playertext}

\vspace{5mm}

\begin{playertext}[title={System instruction given to the basic harness}]
We are not going to tell you the rules of this game---you have to figure them out for yourself.

\textbf{Levels, lives and steps:}
\begin{itemize}[nosep, leftmargin=1.4em, topsep=2pt]
    \item The aim is to reach as high a level as possible. For each level you reach you will be awarded a bonus.
    \item You advance levels by reaching certain states within the game. You will have to figure out what these are.
    \item Within each level, you have a limited number of steps. If you run out of steps, you lose a life. If you lose all your lives, the game is over.
    \item It is also possible to lose a life by reaching certain states within the game.
\end{itemize}

\textbf{Creative mode:}
\begin{itemize}[nosep, leftmargin=1.4em, topsep=2pt]
    \item The game may also have a creative mode. If it does, you can switch into it at any time to experiment safely without losing steps or lives, and it may be necessary to use it to discover the rules without running out of steps. You will know creative mode is available when the state carries a \texttt{mode} field and a toggle action appears in your \texttt{legal\_actions}.
\end{itemize}

\textbf{How you play:}
\begin{itemize}[nosep, leftmargin=1.4em, topsep=2pt]
    \item A short TASK DESCRIPTION at the start gives the objective and any special actions (not the rules). Each turn you then receive the current state: a text \texttt{observation} (the rendered screen), \texttt{level}, \texttt{max\_level}, \texttt{lives\_left}, \texttt{steps\_remaining}, the list of \texttt{legal\_actions}, plus \texttt{mode} and a \texttt{transition} message when those apply.
    \item Infer what each action does from how the state changes, and build on what you learn across turns. Reason carefully, then call \texttt{make\_move} with EXACTLY ONE action from \texttt{legal\_actions}.
\end{itemize}
\end{playertext}

\vspace{5mm}

\begin{playertext}[title={First turn given to the basic harness}]
You are now playing this game.

TASK DESCRIPTION (objective + any special actions, NOT the rules):

\texttt{\{description\}}

INITIAL STATE:

\texttt{\{state\}}

Reason about it, then call \texttt{make\_move}.
\end{playertext}

\vspace{5mm}

\begin{playertext}[title={Prompt template given to the agentic harness}]
We are not going to tell you the rules of this game---you have to figure them out for yourself.

\textbf{Levels, lives and steps:}
\begin{itemize}[nosep, leftmargin=1.4em, topsep=2pt]
    \item The aim is to reach as high a level as possible. For each level you reach you will be awarded a bonus.
    \item You advance levels by reaching certain states within the game. You will have to figure out what these are.
    \item Within each level, you have a limited number of steps. If you run out of steps, you lose a life. If you lose all your lives, the game is over.
    \item It is also possible to lose a life by reaching certain states within the game.
\end{itemize}

\textbf{Important: creative mode}\nopagebreak

At nearly any time, you can use a button to switch into ``creative mode'', where you can experiment safely without losing steps or lives.\\
It may be necessary to use creative mode in order to discover the rules of the game without running out of steps.

Call the ``step'' tool with action \texttt{"/"} to enter creative mode.\\
Call the ``step'' tool with action \texttt{"/"} again to return to survival mode.\\
Only submit \texttt{"/"} when it appears in the state's \texttt{actions} list.

\medskip
\textbf{How you play}---use the ``Agent Benchmark API'' MCP tools to drive the game:
\begin{enumerate}[nosep, leftmargin=1.7em, topsep=2pt]
    \item Your game session is ALREADY started for you:\par\noindent
    \texttt{session\_\allowbreak id=\allowbreak "\{session\_\allowbreak id\}"}, \texttt{game="\{game\}"}. You do NOT start or choose a game---you only have the \texttt{\{tools\}} tools, scoped to this one session, and you must pass this \texttt{session\_id} to every call. Your starting state (\texttt{step\_index=\{step\_index\}}) is:

    \texttt{\{state\}}

    \item Each turn, read the current state and reason from these fields: \texttt{observation} (the rendered screen), \texttt{level}, \texttt{max\_level}, \texttt{lives\_left}, \texttt{steps\_remaining}, \texttt{status}, \texttt{done}, the \texttt{actions} list (your legal moves), and \texttt{mode}/\texttt{transition} when present.
    \item Make a move with the ``step'' tool: \texttt{session\_id="\{session\_id\}"}, \texttt{step\_index=<the server's last returned step\_index + 1>}, \texttt{action=<EXACTLY ONE string from the current state's actions list>}. Your first move uses \texttt{step\_index=\{first\_step\}}. A \texttt{step\_index} mismatch is a 409---always step off the server's last returned \texttt{step\_index}. Use the ``get\_session'' tool if you ever need to re-read the current state. Infer what each action does from how the state changes, and build on what you learn across turns. Keep playing, \texttt{\{pace\}}, until the state's \texttt{done} is true (status \texttt{game\_over} or \texttt{completed}).
\end{enumerate}

When the game is done, STOP making moves and write your debrief: the mechanics you discovered, the objective, useful strategies, and remaining uncertainties.
\end{playertext}

\vspace{5mm}

\begin{playertext}[title={Prompt template given to the Prime Agent}]
    We are not going to tell you the rules of this game---you have to figure them out for yourself.

\textbf{Levels, lives and steps:}
\begin{itemize}[nosep, leftmargin=1.4em, topsep=2pt]
    \item The aim is to reach as high a level as possible. For each level you reach you will be awarded a bonus.
    \item You advance levels by reaching certain states within the game. You will have to figure out what these are.
    \item Within each level, you have a limited number of steps. If you run out of steps, you lose a life. If you lose all your lives, the game is over.
    \item It is also possible to lose a life by reaching certain states within the game.
\end{itemize}

\textbf{Important: creative mode}\nopagebreak

At nearly any time, you can use a button to switch into ``creative mode'', where you can experiment safely without losing steps or lives.\\
It may be necessary to use creative mode in order to discover the rules of the game without running out of steps.

Call \texttt{game\_\allowbreak tool.step} with action \texttt{"/"} to enter creative mode.\\
Call \texttt{game\_\allowbreak tool.step} with action \texttt{"/"} again to return to survival mode.\\
Only submit \texttt{"/"} when it appears in the state's \texttt{actions} list.

\medskip
\textbf{How you play}---use the \texttt{game\_\allowbreak tool} Python module (already importable in your IPython kernel) to drive the game:
\begin{enumerate}[nosep, leftmargin=1.7em, topsep=2pt]
    \item Your game session is ALREADY started for you:\par\noindent
    \texttt{session\_\allowbreak id=\allowbreak "\{session\_\allowbreak id\}"}, \texttt{game="\{game\}"}. You do NOT start or choose a game---you only have the \texttt{game\_\allowbreak tool.step} and \texttt{game\_\allowbreak tool.get\_\allowbreak session} functions, scoped to this one session, and you must pass this \texttt{session\_id} to every call. Your starting state (\texttt{step\_index=\{step\_index\}}) is:

    \texttt{\{state\}}

    \item Each turn, read the current state and reason from these fields: \texttt{observation} (the rendered screen), \texttt{level}, \texttt{max\_level}, \texttt{lives\_left}, \texttt{steps\_remaining}, \texttt{status}, \texttt{done}, the \texttt{actions} list (your legal moves), and \texttt{mode}/\texttt{transition} when present.
    \item Make a move with \texttt{game\_\allowbreak tool.step(\allowbreak session\_\allowbreak id="\{session\_\allowbreak id\}", step\_\allowbreak index=<the server's last returned step\_index + 1>, action=<EXACTLY ONE string from the current state's actions list>)}. Your first move uses \texttt{step\_index=\{first\_step\}}. A \texttt{step\_index} mismatch is a 409---always step off the server's last returned \texttt{step\_index}. Use \texttt{game\_\allowbreak tool.get\_\allowbreak session(\allowbreak session\_\allowbreak id="\{session\_\allowbreak id\}")} if you ever need to re-read the current state.
    \item Infer what each action does from how the state changes, and build on what you learn across turns. Keep playing, one move per turn, until the state's \texttt{done} is true (status \texttt{game\_over} or \texttt{completed}).
\end{enumerate}

When the game is done, STOP making moves and write your debrief: the mechanics you discovered, the objective, useful strategies, and remaining uncertainties.

\end{playertext}

\section{Game unbeaten with rules}
\label{app:unbeaten-with-rules}

\begin{figure}[H]
    \centering
    \includegraphics[width=0.62\textwidth]{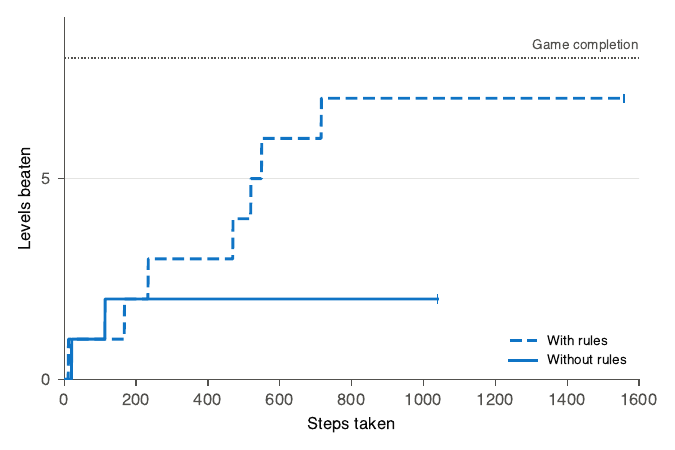}
    \caption{\textbf{Progress on the one game Gemini 3.1 Pro did not beat even when given the rules.} Levels beaten against steps taken. Dashed line is the run with rules supplied, solid line is without rules. Both conditions are Gemini 3.1 Pro. The dotted horizontal line marks game completion (8 levels). With the rules, Gemini beat 7/8 levels across 1,558 steps, without them 2/8 levels across 1,039 steps.}
    \label{fig:unbeaten}
\end{figure}

One game resisted Gemini 3.1 Pro even when the ground truth rules were supplied~(\cref{sec:results}). \cref{fig:unbeaten} compares its progress with and without rules given.

\section{Performance on ARC-like games}
\label{app:arc-results}

\begin{figure}
    \centering
    \widefig{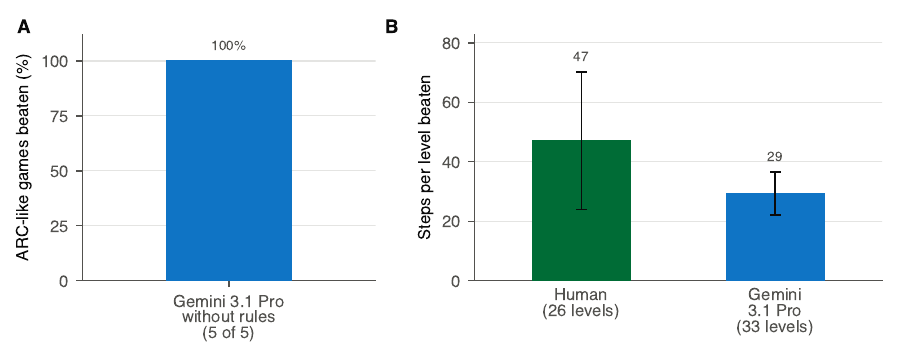}
    \caption{\textbf{Performance on the five ARC-like games.} (\textbf{A}) Games beaten by Gemini 3.1 Pro without rules. (\textbf{B}) Mean steps taken to beat a level by humans versus Gemini 3.1 Pro. Every level beaten by the best human play or the best Gemini 3.1 Pro run on each of the five games contributes one observation, pooled across games. Error bars are 95\% confidence intervals. Gemini 3.1 Pro runs are pooled from the basic harness and the PRO-LONG agentic harness.}
    \label{fig:arc-results}
\end{figure}

ARC-AGI-3 (2026)~\cite{arc_prize_foundation_arc-agi-3_2026} is the interactive installment of the ARC-AGI (Abstraction and Reasoning Corpus for Artificial General Intelligence) series, created by François Chollet and maintained by the ARC Prize Foundation. It is designed to measure skill-acquisition efficiency and adaptation to novel environments. ARC-AGI-3 presents games without instructions, requiring the player to explore, plan, and discover the rules by acting. However, the games are conceptually simpler and require less experimentation to uncover the winning principles. The ARC benchmark evaluates step efficiency as a key metric, rewarding agents that reach a solution in fewer actions. It also represents its environments visually in a two-dimensional grid. To gauge how much of the difficulty of ARC-AGI-3 comes from the reliance on visual and spatial reasoning, we ported five public ARC-AGI-3 games into our text-based framework. Gemini 3.1 Pro cleared every level of all five, and the human step-efficiency advantage reported on the originals did not reproduce (\cref{fig:arc-results}).

We created five other games (not included in the benchmark) intended to mimic as closely as possible the logic of five public ARC-AGI-3 games while being expressed in short strings rather than grids. It was not possible to make the game logic precisely isomorphic to the ARC-AGI-3 games, because some of the puzzle logic in ARC is intrinsically spatial. Nevertheless, these new games provided a crude view on how challenging the ARC puzzles would be if they were not confounded with perception and spatial reasoning. 

Gemini 3.1 Pro cleared every level of all five ARC-like games (\cref{fig:arc-results}A). This result is consistent with the difficulty of the originals depending partly on other aspects of the ARC-AGI-3 environment, such as visual perception. On these ported games we also did not reproduce the human step-efficiency advantage of the original ARC-AGI-3 set. Gemini's mean efficiency was slightly better at 29 steps per level against 47 for humans (\cref{fig:arc-results}B). The medians were nearly identical: 23 steps for humans and 24 for Gemini.

\FloatBarrier
\section{Runs stopped at cost or time caps}
\label{app:cap-stopped-runs}

\cref{tab:cap-stopped-runs} reports the number of runs stopped by each condition's per-game cost or wall-clock cap.

\begin{table}[ht]
\centering
\small
\begin{tabular}{llr}
\toprule
\textbf{Harness} & \textbf{Model} & \textbf{Cap-stopped runs} \\
\midrule
basic harness & Claude Opus 5 & 1 \\
basic harness & Gemini 3.1 Pro Preview & 4 \\
basic harness & GPT-5.5 & 3 \\
basic harness & Kimi K3 & 6 \\
basic harness & GLM-5.2 & 2 \\
basic harness & DeepSeek V4 Pro Preview & 0 \\
basic harness & DeepSeek V4 Flash 0731 & 0 \\
basic harness & Qwen3.6 27B & 0 \\
agentic harness (Prime Agent) & Claude Opus 5 & 0 \\
agentic harness (Claude Code) & Claude Fable 5 & 5 \\
agentic harness (Codex) & GPT-5.6 Sol & 1 \\
agentic harness (Kimi Code) & Kimi K3 & 1 \\
agentic harness (PRO-LONG) & Gemini 3.1 Pro Preview & 0 \\
\bottomrule
\end{tabular}
\caption{Runs stopped by a per-game cost or wall-clock cap.}
\label{tab:cap-stopped-runs}
\end{table}

\end{document}